%% file: ijcai22.tex
\documentclass{article}
\usepackage{ijcai22}

\usepackage{times}
\usepackage{soul}
\usepackage{url}
\usepackage[hidelinks]{hyperref}
\usepackage[utf8]{inputenc}
\usepackage[small]{caption}
\usepackage{graphicx}
\usepackage{amsmath}
\usepackage{amsthm}
\usepackage{booktabs}
\usepackage{algorithm}
\usepackage{algorithmic}
\usepackage{float}
\usepackage{mathtools}
\usepackage{multirow}
\usepackage{color}
\usepackage{CJKutf8}
\usepackage[T1]{fontenc}
\title{TURNER: The Uncertainty-based Retrieval Framework for Chinese NER}

\author{
Zhichao Geng
\and
Hang Yan\and
Zhangyue Yin\and
Chenxin An\and
Xipeng Qiu\thanks{\ \ Corresponding author}\and
Xuanjing Huang
\affiliations
School of Computer Science, Fudan University\\
Key Laboratory of Intelligent Information Processing, Fudan University \\
\emails
\{zcgeng20,hyan19,xpqiu,xjhuang\}@fudan.edu.cn
}

\begin{document}

\maketitle

\begin{abstract}
Chinese NER is a difficult undertaking due to the ambiguity of Chinese characters and the absence of word boundaries.
Previous work on Chinese NER focus on lexicon-based methods to introduce boundary information and reduce out-of-vocabulary (OOV) cases during prediction.
However, it is expensive to obtain and dynamically maintain high-quality lexicons in specific domains, which motivates us to utilize more general knowledge resources, e.g., search engines.
In this paper, we propose TURNER: \textbf{T}he \textbf{U}ncertainty-based \textbf{R}etrieval framework for Chinese \textbf{NER}. 
The idea behind TURNER is to imitate human behavior: we frequently retrieve auxiliary knowledge as assistance when encountering an unknown or uncertain entity.
To improve the efficiency and effectiveness of retrieval, we first propose two types of uncertainty sampling methods for selecting the most ambiguous entity-level uncertain components of the input text.
Then, the Knowledge Fusion Model re-predict the uncertain samples by combining retrieved knowledge.
Experiments on four benchmark datasets demonstrate TURNER's effectiveness. TURNER outperforms existing lexicon-based approaches and achieves the new SOTA.

\end{abstract}

\section{Introduction}

Named Entity Recognition (NER) plays an essential part in natural language processing.
The performance of the NER task has improved dramatically as a result of the recent advances in deep learning and pre-trained models.
In comparison to English NER, Chinese NER is more challenging due to the character ambiguity and absence of word boundaries.

\begin{figure}[t]
\centering
\includegraphics[page=1,width=0.47\textwidth]{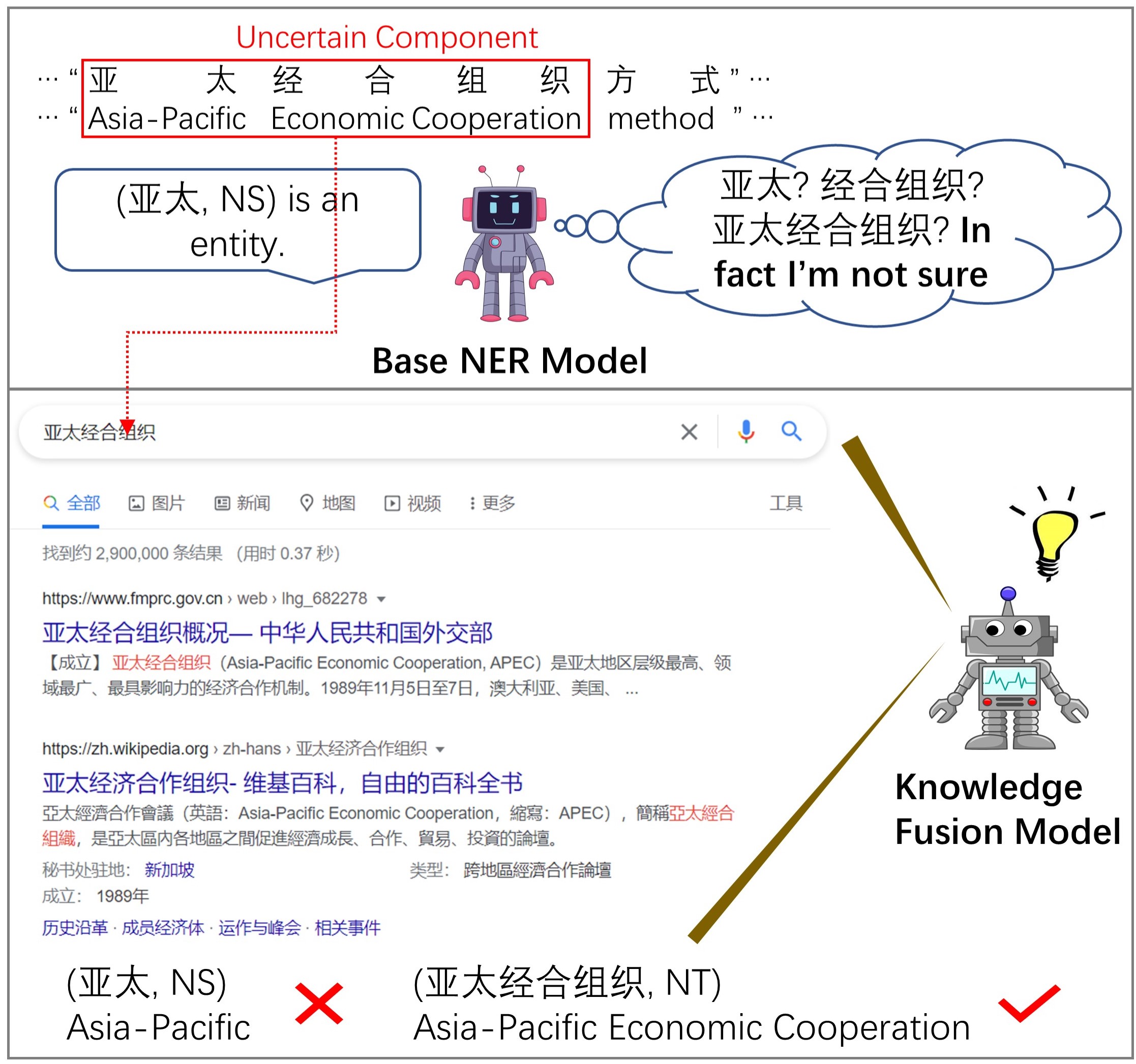}
\caption{A case of our Uncertainty-based Retrieval framework. The base NER model identifies "Asia-Pacific" as an NS entity, but it is uncertain whether "Asia-Pacific Economic Cooperation" is an NT entity. The Knowledge Fusion Model accurately predicts by retrieving knowledge using the query "Asia-Pacific Economic Cooperation" with the search engine.}
\label{pic:intro}
\end{figure}

In recent years, lexicon-based approaches have dominated the Chinese NER field~\cite{YueZhang2018ChineseNU,li2020flat,liu-etal-2021-lexicon,wang2021dylex}.
Due to their ability to reduce OOV occurrence during prediction, lexicon-based approaches have become the necessary element for obtaining SOTA results in Chinese NER.
These approaches begin by matching the input sentence with the lexicon and then incorporating the matched items' boundary and semantic information into the model via a variety of methods.
However, the lexicon-based approaches have several drawbacks.
First, model performance is highly dependent on the quality of the lexicon, yet obtaining and dynamically maintaining a high-quality lexicon in a particular application scenario is costly.
Second, existing methods are incapable of utilizing dynamically changing lexicons while introducing semantic knowledge and vice versa.
A line of previous work~\cite{li2020flat,liu-etal-2021-lexicon} utilizes word embeddings to represent matched items, but updating the lexicon requires retraining the model.
Another tributary of previous work~\cite{wang2021dylex} leverages only the word boundary information, which enables the usage of a dynamically updated lexicon, but this method discards the semantic information of words.

Considering the limits of lexicon-based methods, retrieval methods, e.g. retrieving with search engines, are more generic resources of of knowledge in the NER task.
The retrieved knowledge can help eliminate ambiguity in the input text and enhance performance.
Additionally, the retrieved knowledge is presented as text, allowing the query-knowledge base to be dynamically updated.
However, existing retrieval methods~\cite{torisawa2007exploiting,wang-etal-2021-improving} in the NER task retrieve the knowledge using traversal strategies, which is inefficient, especially in the case of online retrieval.
Moreover, existing pre-trained models are sufficiently powerful to handle the majority of NER cases.
Retrieving knowledge for simple samples that the model could have dealt with correctly is ineffective.

We notice that in Chinese NER, the model can produce diverse predictions about some input components, i.e., the model is uncertain about these components.
Furthermore, our statistics demonstrate that the uncertain components are the bottleneck of performance.
Therefore, retrieving knowledge for the uncertain components can provide the maximum support for the prediction.
In this paper, we propose TURNER: \textbf{T}he \textbf{U}ncertainty \textbf{R}etrieval framework for Chinese \textbf{NER}. 
By leveraging Monte-Carlo Dropout~\cite{YarinGal2016DropoutAA} or parsing the top-$K$ label sequences predicted by the base NER model, we can obtain the entity-level uncertain components with a high degree of complexity and uncertainty for the base NER model.
Then by querying on the uncertain components, knowledge can be retrieved efficiently and effectively, eliminating uncertainty, as shown in Figure~\ref{pic:intro}.
Afterward, the Knowledge Fusion Model re-predicts the uncertain samples by integrating the original input text, the retrieved knowledge, and the output of the base NER model.
We conduct experiments on 4 Chinese NER datasets to evaluate the effectiveness of our method.
The results illustrate TURNER significantly outperforms the base NER model on all datasets, boosting the F1 score by 1.1 on average.
TURNER also outperforms existing lexicon-based methods on Chinese NER, achieving new SOTA results.
To the best of our knowledge, we are the first to combine uncertainty sampling and  retrieval methods in the NER task.

\section{Related Work and Background}
\input{content/topk_how}
\subsection{Chinese Named Entity Recognition}
With the advent of deep learning, neural networks have become a dominant solution to the NER task~\cite{XuezheMa2016EndtoendSL}. 
In recent years, lexicon-based methods have become widespread for Chinese NER.
Zhang and Yang~\shortcite{YueZhang2018ChineseNU} propose lattice LSTM to encode both lexicon information and character information.
Li~\shortcite{li2020flat} propose the Flat-Lattice Transformer as a way to incorporate lexicon information into the transformer model without any loss of information.
Liu~\shortcite{liu-etal-2021-lexicon} make use of the BERT Adapter~\cite{NeilHoulsby2019ParameterEfficientTL} to introduce lexicon information to the bottom layers of BERT.
Wang~\shortcite{wang2021dylex} propose DyLex, which utilizes only the boundary information of words, allowing models to use dynamically updating lexicons.

\subsection{Retrieval Methods for Neural Models}
Retrieving knowledge from the database to enhance neural models has been applied in several NLP tasks~\cite{XipengQiu2014AutomaticCE,TatsunoriBHashimoto2018ARF,gu2018search}. 
They either use the retrieved text to guide the generation process or expand the training set automatically.
In the NER task, some works introduce knowledge from the knowledge graph or Wikipedia~\cite{torisawa2007exploiting,he2020knowledge}.
They retrieve knowledge by traversing matches all possible sub-sequences, then encode the knowledge to incorporate it into the model.
Wang~\shortcite{wang-etal-2021-improving} chunk the sentence into sub-sentences based on punctuation, then utilizes the sub-sentences as queries in the search engine to retrieve external context.
The knowledge is chosen from the retrieved items that have the highest association with the original text.

\subsection{Uncertainty}
The predictive result in the classification task can be uncertain even with a high softmax output~\cite{YarinGal2016DropoutAA}.
Monte-Carlo Dropout~\cite{YarinGal2016DropoutAA} (MC Dropout),  a Bayesian approximation of the Gaussian process, is a general approach to estimate uncertainty.
MC Dropout keeps dropout active in the prediction phase and evaluates the uncertainty according to the difference of several forward passes.
In the sequence labeling task, Gui~\shortcite{TaoGui2020UncertaintyAwareLR} employ MC Dropout to create draft labels and locate the uncertain labels. 
Furthermore, they utilize two-stream attention to model long-term dependencies between draft labels and refine them, thereby getting rid of Viterbi decoding.

\section{Approach}
In this section, we will first introduce our uncertainty sampling methods for sequence labeling NER models, followed by our framework for Chinese NER utilizing the uncertainty.

\subsection{Uncertainty Sampling and Retrieving}

\subsubsection{Sampling Method}
\label{sec:samp_method}
\paragraph{MC Dropout}
MC Dropout~\cite{YarinGal2016DropoutAA} is a general approach to obtain the uncertain components.
Given an input sequence for the sequence labeling NER model, we first obtain the model's prediction as the provisional result.
Then, we utilize MC dropout to keep dropout active and generate $K$ candidate label sequences with Viterbi decoding in $K$ forward passes.
As illustrated in Table~\ref{tab:topkhow}, the difference between the predicted entity set of each candidate label sequence and the provisional result can be considered uncertain entities as the model predict them under mild perturbation.
The uncertain components can then be obtained by merging all adjacent and overlapping uncertain entities.
We sample uncertain entities rather than position-level labels because querying at the entity level improves retrieval precision.

\paragraph{Top-$K$ Label Sequences}
MC dropout demands more GPU resources as it requires multiple forward passes.
As a result, we also give an alternative strategy for sampling uncertainty in NER tasks.
Given the label probability distribution of an input sequence, we can obtain $K$ legal label sequences with top-$K$ scores based on the variant of the Viterbi algorithm~\cite{brown2010decoding}.
Previous works rerank the top-$K$ label sequences to enhance the performance~\cite{JieYang2017NeuralRF}, but we can also sample the uncertainty from these sequences.
We adopt the top-1 label sequence as the provisional result and other label sequences as the candidate label sequences.
Afterward, we can sample the uncertainty as described before.
Additionally, we observe that when the model is highly confident about the prediction, the difference between the prediction result and the top-$K$ label sequences is minor, such as altering a single $O$ label to an $S$ type label.
Because this will minimize the loss of scores when there is no competing entity in candidates.
Therefore, we filter out candidates that differ in a only single position from the provisional result to minimize the retrieval times, considering most Chinese entity names are longer than one character.

\subsubsection{Preliminary Results}
To verify the importance of the uncertainty components, we conduct an investigation on the test set of four benchmark Chinese NER datasets utilizing BERT as the base NER model.\footnote{More details of our settings can be found in Section~\ref{sec:exp}.}
We generate eight candidate label sequences using MC Dropout, and the results are shown in Table~\ref{tab:topkmc}.
\input{content/topk_MC}

The base NER model obtains high oracle F1 scores indicating that it possesses considerable promise.
Even while the base NER model achieves an impressive level of accuracy on the confident components, the significant gap between ACC$_{certain}$ and ACC$_{uncertain}$ reveals that the uncertain components are real hard samples and become bottlenecks for performance.
Therefore, by querying about uncertain components, the desired knowledge can be retrieved efficiently and effectively.
Additionally, the proportion of uncertain components is inversely proportional to the model's performance across datasets, and the number of retrievals is significantly less than the number of test samples in all datasets.

\subsubsection{Retrieving}
The search engine is a general resource for acquiring supplemental knowledge for the uncertain components.
By querying the search engine, we can obtain the context related to the uncertain components as auxiliary knowledge to support the prediction.
Similarly, if available, we can also retrieve knowledge offline from the in-domain encyclopedias or knowledge graphs using retrieval algorithms such as BM25.
When search engines are adopted, the cost of retrieval can be significant.
However, for a particular application scenario, the distribution of entities is usually concentrated. 
Therefore, as more query-knowledge pairs are collected, the number of new requests will continue to decrease.

\subsection{Framework}

\begin{figure}[htb]
\centering
\includegraphics[page=1,width=0.47\textwidth]{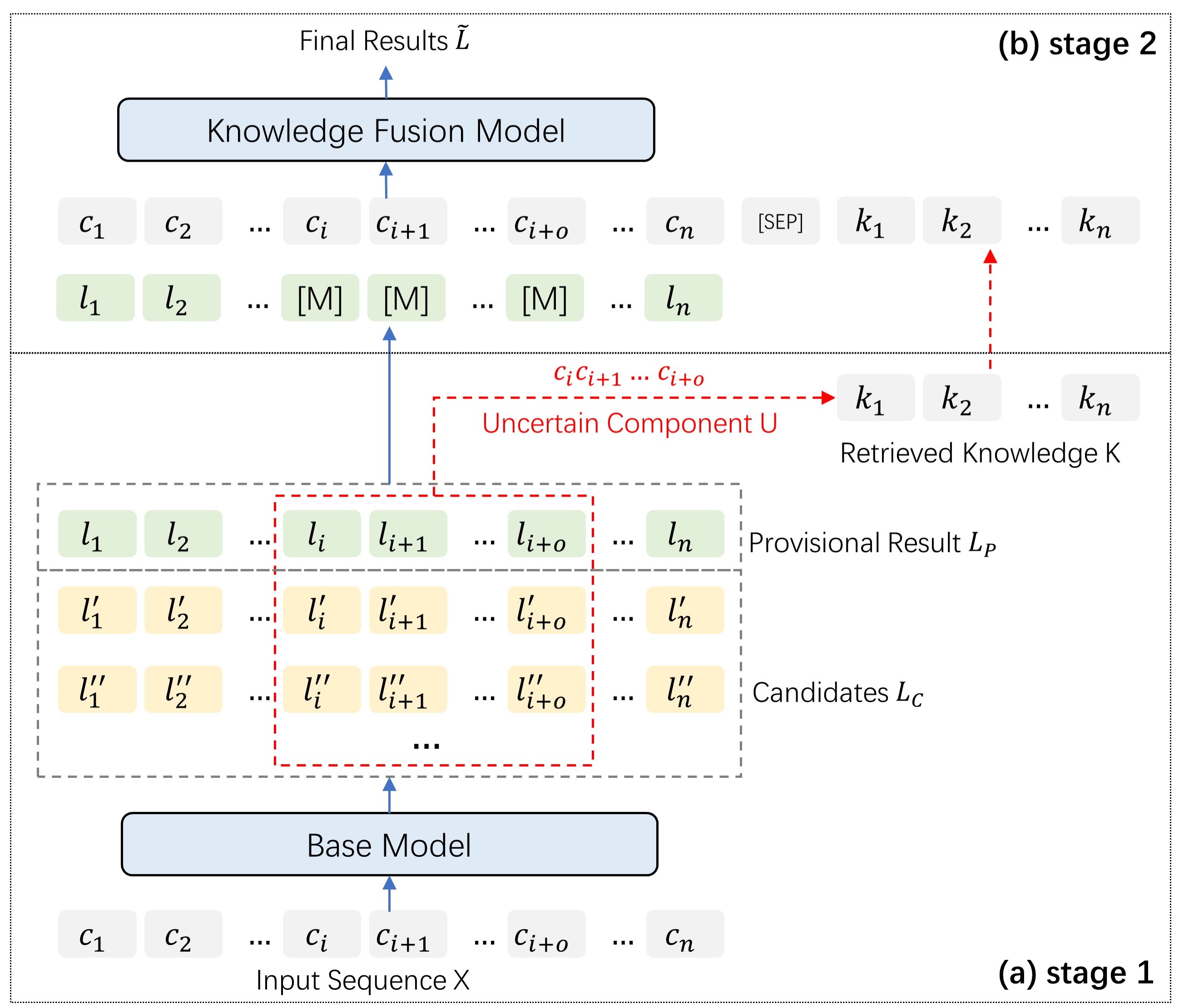}
\caption{The architecture of TURNER. (a) The base NER model makes predictions on the input sequence. This enables TURNER to identify uncertain components and then retrieve pertinent knowledge. (b) The Knowledge Fusion Model produces predictions based on the retrieved knowledge and the provisional result.}
\label{pic:model}
\end{figure}

In this part, we will present the architecture of TURNER.
As illustrated in Figure~\ref{pic:model}, TURNER solves the uncertain components via a two-stage pipeline.
We employ a base NER model to acquire provisional results and uncertain components in the first stage and another Knowledge Fusion Model to re-predict the uncertain samples with retrieved knowledge in the second stage.
The parameters of the Knowledge Fusion Model and the base NER model are independent of each other.

\subsubsection{Stage One: Provisional Result and Uncertainty Sampling}
In the first stage, we use a base NER model to predict on the input sequence to get provisional results and uncertain components.
The base NER model can be any sequence labeling NER model, such as BERT+MLP.

Formally, given input sequence $X=[c_{1},c_{2},...,c_{n}]$, we apply the base NER model to obtain the provisional results $L_{P}$ and candidate label sequences $L_{C}$ with dropout active.
If there is no uncertain component in all candidates, the base NER model is highly confident about the current sample, then $L_{P}$ is the final result.
Otherwise, we take $L_{P}$ as the provisional result and obtain the uncertain component $U=[c_i,c_{i+1},...,c_{i+o}]$ using the method in Section~\ref{sec:samp_method}.
Afterward, we can utilize U as query to retrieve the knowledge text $K=[k_{1},k_{2},...,k_{m}]$.
If there are multiple non-continuous uncertain components in one input sequence, we retrieve the knowledge separately and process them independently in stage two.

\subsubsection{Stage Two: Knowledge Fusion Prediction}
In the second stage, We employ $K$ as auxiliary knowledge to re-predict the sample with uncertain component $U$ using the Transformer-based Knowledge Fusion Model.
To introduce the retrieved knowledge without losing any information, we concatenate $K$ behind $X$ to obtain the knowledge-enhanced input sequence $\tilde{X}=[c_{1},c_{2},...,c_{n},[SEP],k_{1},k_{2},...,k_{m}]$.

Considering the base NER model's excellent accuracy in the confident components, we incorporate the label information of the provisional result to support the Knowledge Fusion Model.
To avoid confusion or misleading, we mask the uncertain labels in the provisional result and obtain the modified provisional result $\tilde{L_{P}}$ as follows:
\begin{align}
&\tilde{l_{i}}=\begin{cases}l_{i}~~~~~~~~~~\text{if}~i\le n~\text{and}~c_{i}\notin U
 \\ [mask]\quad\text{if}~c_{i}\in U
 \\ [pad]~~~~~~~~\text{if}~i>n
 \end{cases},\\
&\tilde{L_{P}}=[\tilde{l_{1}},\tilde{l_{2}},...\tilde{l_{n}},\tilde{l_{n+1}},...,\tilde{l_{n+m+1}}],
\end{align}
where $l_{i}$ is the i-th label of the provisional result and $\tilde{l_{i}}$ is the i-th label in $\tilde{L_{P}}$.
Afterward, we incorporate $\tilde{X}$ and $\tilde{L_{P}}$ using the Transformer-based Knowledge Fusion Model to obtain the knowledge-enhanced probability distribution $\tilde{D}$:
\begin{align}
&H_{\tilde{L_{P}}}=\text{Label Embedding}(\tilde{L_{P}}),\\
&H_{\tilde{X}}=\text{Character Embedding}(\tilde{X}),\\
&\tilde{D}=\text{Transformer Encoder}(H_{\tilde{L_{P}}}+H_{\tilde{X}}),
\end{align}
where label embedding and character embedding are trainable parameters.
The self-attention mechanism ensures that all information is fully integrated and the knowledge is efficiently utilized.
Ultimately, we decode $\tilde{D}$ using the Viterbi algorithm to obtain the final result $\tilde{L}$.
If there are multiple uncertain components in one input sequence, we process them separately in the second stage and sum all obtained $\tilde{D}$ before Viterbi decoding.

\subsection{Training of TURNER}
Since the parameters of the base NER model and the Knowledge Fusion Model are independent, the two-stage training procedure can be undertaken independently.
The training of the base NER model is trivial, and we will focus on the training of the Knowledge Fusion Model.

To minimize the gap between training and prediction, the uncertainty of training data must be sampled.
Therefore, we conduct N-fold jackknifing to divide the training data into N equal parts.
In each step, we train the base NER model with N-1 pieces of data and then use the base NER model to sample uncertainty for the remaining one piece of data.
Additionally, we conduct data augmentation by leveraging various checkpoints of the base NER model when generating training data for the second stage.
For a given training sample, we append the sampled components of all checkpoints sequentially and discard the components that are significantly overlapped with existing uncertain components.

However, the data augmentation strategy will result in the repetition of some training samples, which may lead to overfitting on several input sequences.
Therefore, we introduce the position-wised weighted average for the cross-entropy loss.
By decreasing the weight assigned to the confident component, the model can concentrate on the uncertain components rather than overfitting the simple cases.
Formally, We calculate the loss $\mathcal{L}$ as follows:
\begin{align}
&\lambda _{i}=\begin{cases}1~~~~~~\text{if}~c_{i}\in U
 \\ \alpha ~~~~~~\text{if}~c_{i}\notin U
\end{cases},\\
&\mathcal{L}=\frac{\sum_{i}^{1\le i\le n}\cdot\lambda _{i}\cdot loss_{i}}{\sum_{i}^{1\le i\le n}\cdot\lambda _{i}},
\end{align}
where $\lambda_{i}$ is the weight coefficient at position $i$; $loss_{i}$ is the cross-entropy loss at position $i$; $\alpha$ is a hyper parameter ranges $[0,1]$.
And we do not calculate loss for the knowledge text.

\section{Experiments}
\label{sec:exp}
We conduct comprehensive experiments to evaluate the effectiveness of TURNER.
We use the $BIESO$ label set and employ the standard F1-score as the evaluation metric.
All experiments are conducted on 8 GeForce RTX 3090.

\subsection{Datasets}
We conduct experiments on four benchmark Chinese NER datasets to evaluate TURNER, including MSRA~\cite{levow2006third}, Ontonotes 4.0~\cite{weischedel2011ontonotes}, Resume~\cite{YueZhang2018ChineseNU} and Weibo~\cite{peng2015named}.
MSRA and Ontonotes are annotated using data from the news domain; Weibo is annotated from the Internet blogs; Resume is annotated from the resumes. 
We follow the same train, dev, test split of the official version and previous work~\cite{li2020flat}.
The statistical information of each dataset is shown in Table~\ref{tab:datasets}.
\input{content/datasets}

\subsection{Settings}
\paragraph{Models and Training}
We implement a strong base NER model using BERT\footnote{\url{https://huggingface.co/hfl/chinese-bert-wwm}}+MLP in the first stage of TURNER.
We do not use CRF because we observe that the same performance can be obtained by relying just on the Viterbi algorithm to constrain that there is no illegal transition in the predictive result.
The Knowledge Fusion Model is also initialized using BERT.
We search the learning rate in \{2e-5, 3e-5\} and the weight coefficient $\alpha$ in \{0.1, 1\}, and other hyperparameters we used are listed in Appendix A.
We use the validation set to select the best checkpoint and report its performance on the test set.

\paragraph{Uncertainty and Retrieval}
During prediction, We generate eight candidates for MC Dropout and four candidates for the top-$K$ method.
To generate training data for the second stage of TURNER, we use MC Dropout for uncertainty sampling.
The statistical results of the uncertainty sampling on four datasets are listed in Appendix B.
For retrieving, we use {\it Baidu}, which is a widely used Chinese search engine. 
We prioritize the retrieved items belonging to the encyclopedia category and do not change the order of other retrieved items.
Furthermore, each retrieved item is limited to a title and up to 50 characters of content, and all items are concatenated in order.
The retrieved knowledge is limited to 400 characters.

\subsection{Overall Performance}
Table~\ref{tab:main} shows the overall performance of TURNER in four benchmark datasets of Chinese NER.
\input{content/mainres}
In the top part of table we show the performance of lexicon-free baselines, including BiLSTM+CRF~\cite{huang2015bidirectional}, TENER~\cite{yan2019tener} and ERNIE~\cite{sun2019ernie}.
In the second part of table we provide several latest strong lexicon-based baselines, including Lattice-LSTM~\cite{YueZhang2018ChineseNU}, FLAT+BERT~\cite{li2020flat}, LEBERT~\cite{liu-etal-2021-lexicon}, DyLex+BERT~\cite{wang2021dylex}.
In the bottom part, we show the performance of TURNER with two uncertainty sampling methods.
The results demonstrate that TURNER significantly enhances the base NER model's performance, boosting the average F1 score by 1.1 with both two uncertainty sampling methods and achieving new SOTA results.
Although lexicon-based baselines are obviously superior to lexicon-free baselines, TURNER outperforms existing lexicon-based methods in each dataset, proving the potential of lexicon-free methods in Chinese NER.

TURNER makes more substantial advancements in MSRA and Ontonotes, which belong to the news domain, because entities referenced in the news are more likely to be retrieved by search engines.
The improvement on Weibo is relatively limited, as many entities are usernames or nicknames, making it more challenging to retrieve related information using the search engine.
It can be inferred that the effectiveness of TURNER is directly proportional to the relevance of the retrieved knowledge.

\subsection{Ablation Study}
We conduct ablation experiments to verify the contribution of each component of TURNER.
The settings of the ablation experiment are shown as follows:
\textbf{(a) W/o retrieve:}
We do not retrieve knowledge for the uncertain components, i.e., the input for the Knowledge Fusion Model $\tilde{X}$ equals $X$. 
\textbf{(b) W/o uncertainty:}
We abandon the two-stage architecture and do not sample the uncertainty. 
As a result, we use sentence-level queries to retrieve knowledge for each sample like Wang\shortcite{wang-etal-2021-improving}. 
The knowledge is also concatenated behind the original input text.
\textbf{(c) W/o LE:}
We do not use the label embedding to provide the provisional results to the Knowledge Fusion Model.
\textbf{(c) W/o WL:}
We use the unweighted average method to calculate the loss for the Knowledge Fusion Model, which is equivalent to setting $\alpha$ to 1.

\input{content/ablation}

As shown in Table~\ref{tab:ablation}, the two-stage uncertain samples re-predicting architecture can be regarded as a particular way of model ensembling, which can boost the performance a bit without knowledge.
Without uncertainty sampling, the contribution of context retrieved via sentence-level queries is limited.
By combining uncertainty sampling and retrieval, TURNER can boost the performance to the greatest extent.
The label embedding and weighted-average loss also contribute much to the Knowledge Fusion Model.

\section{Discussion}
\subsection{TURNER With Offline Retrieval}

The time cost of retrieval from search engines is still too high in some application scenarios.
As a result, we conduct experiments with an offline knowledge graph to test TURNER's generalization capabilities.
We use the {\it ownthink}\footnote{\url{https://www.ownthink.com/knowledge.html}} knowledge graph, which contains more than 140 million $(Subject,Predicate,Object)$ triplets about general knowledge.
For each subject in the knowledge graph, we merge all related triplets containing it to generate a descriptive document~\footnote{More details can be found in Appendix C.}.
Then, for each query, we use the BM25 algorithm to retrieve the three most relevant descriptive documents as auxiliary knowledge in TURNER.
\input{content/offline}

As illustrated in Table~\ref{tab:offline}, TURNER with offline retrieval outperforms existing SOTA lexicon-based methods in MSRA, Ontonotes, and Resume, which can be considered in-domain datasets for the knowledge graph.
Because many internet phrases are out-of-domain for the knowledge graph, the offline retrieval does not operate effectively in Weibo.
The results indicate the generalization ability of TURNER, which can work with various knowledge bases.
The performance of offline retrieval is not as good as the search engines because search engines have broader knowledge and more precise retrieval.

\subsection{MC Dropout vs. Top-$K$ Label Sequences}
Because the two uncertainty sampling approaches are on par in performance, we undertake a more detailed analysis to compare them.
We use two metrics to measure the quality of the uncertainty sampling process:
(a) \textbf{SAR}: since most candidates are duplicate and will be filtered out, we use Sampling Acceptance Ratio (SAR) to denote the ratio of candidates that can be utilized in uncertainty sampling rather than being filtered out;
(b) \textbf{VSR}: we use Valuable Sampling Ratio (VSR) to denote the ratio of candidates that yield a better F1 score than the provisional results after the filter.
The higher the VSR, the better the uncertainty sampling recall ability.
And the VSR/ASR ratio reflects how much noise is present in the sampling results.
As illustrated in Table~\ref{tab:compare}, the top-$K$ method has stonger ability to recognize uncertain components and yield better oracle scores, but there is also more noise, resulting in more retrieval times.
\input{content/compare}

From the perspective of computational efficiency, the additional GPU cost of two sampling methods can be calculated as follows:
\begin{align}
&COST_{\textrm{MC}}=O((k+\beta_{1} (1+\gamma )^{2})\cdot C), \\
&COST_{\textrm{Top-}K}=O(\beta_{2} (1+\gamma )^{2}\cdot C),
\end{align}
where k is the number of candidates, $\beta$ is the retrieval ratio,  $\gamma$ is is the ratio of the length of the knowledge text to the original input text, and $C$ is the cost of the base NER model, i.e., BERT+MLP.
MC Dropout demands more GPU resources in the first stage, while the top-$K$ method results in more retrieval times and consumes more GPU resources in the second stage because it samples more uncertain components.
The ultimate choice should be based on actual details.

\section{Conclusion and Future Work}
% \section{Conclusion}
In this paper, we propose TURNER, a framework for Chinese NER that integrates uncertainty sampling and knowledge retrieval.
TURNER applies MC Dropout or top-$K$ label sequences to identify the uncertain components that lead to ambiguity with a base NER model.
Afterward, the uncertain components can be utilized as queries to retrieve knowledge efficiently and effectively.
Furthermore, the Knowledge Fusion Model re-predict the uncertain samples based on the retrieved knowledge.
Comprehensive experiments on four datasets illustrate TURNER significantly outperforms existing lexicon-based methods and achieves new SOTA results. 
% \section{Future Work}

For existing NLP technologies, it is impossible to dynamically maintain all required knowledge in neural models.
The retrieval methods utilizing external knowledge bases can be solutions to this issue, while our uncertainty-based retrieval paradigm makes retrieval methods more effective and efficient.
In the future, we will try to broadcast the uncertainty-based retrieval paradigm to more NLP tasks and applications.

\clearpage
\bibliographystyle{named}
\bibliography{ijcai22}

\clearpage
 \input{content/appendix}

\end{document}

%% file: content/topk_how.tex
\begin{CJK}{UTF8}{gbsn}
\renewcommand\arraystretch{1.1}
 \begin{table*}[t]
 \scriptsize
  \centering
  \tabcolsep0.14 in
    \begin{tabular}{ccccccccccc|c}
    \toprule
    & " & \multicolumn{2}{c}{Asia-Pacific} & \multicolumn{4}{c}{Economic Cooperation} & \multicolumn{2}{c}{method} & " &\textbf{\multirow{2}{*}{Predicted Entities}} \\
    & " & 亚 & 太 & 经 & 合 & 组 & 织 & 方 & 式 &  "   & \\
    \midrule
    Provisional Result & O  &  B-NS &  E-NS & O & O & O & O & O & O & O  & (亚太,NS) \\
    Candidate 1 & O  &  B-NS &  E-NS & B-NT & I-NT & I-NT & E-NT & O & O & O  & (亚太, NS), (经合组织, NT) \\
    Candidate 2 & O  &  B-NS &  E-NS & O & O & O & O & O & O & O  & (亚太,NS) \\
    Candidate 3 & O  &  B-NT &  I-NT & I-NT & I-NT & I-NT & E-NT & O & O & O  & (亚太经合组织, NT) \\
    Candidate 4 & O  &  B-NS &  E-NS & O & O & B-NT & E-NT & O & O & O  & (亚太,NS), (组织, NT) \\
    \midrule
    \multicolumn{8}{c}{Uncertain Entities: 亚太, 经合组织, 亚太经合组织,  组织}&\multicolumn{4}{c}{Uncertain Component: 亚太经合组织}\\
    \bottomrule
    \end{tabular}
    
\caption{An example of obtaining the uncertain component of an input sequence. The top part shows the input sequence. The middle part lists the candidate label sequences and correlated predicted entities. The bottom part contains uncertain entities and the uncertain component. The candidate label sequences can be obtained from MC Dropout or Top-$K$ label sequences.}
\label{tab:topkhow}
\end{table*}
\end{CJK}

%% file: content/topk_MC.tex
\renewcommand\arraystretch{1.2}
 \begin{table}[h]
%  \scriptsize
\centering
\scalebox{0.85}{
  \tabcolsep0.15 in
    \begin{tabular}{lcccc}
    \toprule
     & MSRA & Onto & Weibo & Resume \\
     \midrule
    F1 Score & 95.92 & 81.54 & 70.11 & 95.78\\
    Oracle F1 & 98.09 & 88.21 & 81.20 & 97.50 \\
    ACC$_{uncertain}$ & 69.51 & 58.02 & 51.08 & 50.93 \\
    ACC$_{certain}$ & 99.69 & 98.12 & 98.01 & 98.64 \\
    Avg UC Num & 0.089 & 0.345 & 0.507 & 0.111 \\
    Avg UC Length & 4.48 & 3.74 & 3.03 & 11.1\\
    \bottomrule
    \end{tabular}
}
\caption{
The statistical results of the uncertain components.
"F1 Score" denotes the F1 score of the base NER model on the test dataset.
"Oracle F1" denotes the F1 score obtained by the base NER model when all uncertain labels are corrected.
"ACC$_{uncertain}$" and "ACC$_{certain}$" denote the position-wise label accuracy of the prediction results for the uncertain components and the confident components, respectively. 
"Avg UC Num" denotes the average number of uncertain components in each input sequence. "Avg UC Length" denotes the average number of tokens in each uncertain component. }
\label{tab:topkmc}
\end{table}

%% file: content/datasets.tex
\begin{table}[htbp]
  \centering
\scalebox{0.95}{
    \begin{tabular}{ccccc}
    \toprule
          & \textbf{Type} & \textbf{Train} & \textbf{Validation} & \textbf{Test} \\
          \midrule
    \multirow{2}[0]{*}{\textbf{MSRA}} & Size  & 46364 & -     & 43.65 \\
          & Length$_{avg}$ & 46.8  & -     & 39.5 \\
          \midrule
    \multirow{2}[0]{*}{\textbf{Ontonotes}} & Size  & 15724 & 4301  & 4346 \\
          & Length$_{avg}$ & 31.3  & 46.6  & 47.9 \\
          \midrule
    \multirow{2}[0]{*}{\textbf{Weibo}} & Size  & 1350  & 270   & 270 \\
          & Length$_{avg}$ & 54.7  & 53.7  & 55 \\
          \midrule
    \multirow{2}[0]{*}{\textbf{Resume}} & Size  & 3821  & 463   & 477 \\
          & Length$_{avg}$ & 32.5  & 30    & 31.7 \\
          \bottomrule
    \end{tabular}
    }
    \caption{Dataset statistics. "Size" denotes the number of samples in the sub-set. "Length$_{avg}$" denotes the average length of samples.}
  \label{tab:datasets}
\end{table}

%% file: content/mainres.tex
\begin{table}[htbp]
\centering
\scalebox{0.9}{
    \begin{tabular}{lcccc}
    \toprule
  & \textbf{MSRA}  & \textbf{Ontonotes} & \textbf{Weibo} & \textbf{Resume} \\
    \midrule
    BiLSTM+CRF & 91.87 & 71.81  & 56.75 & 94.41\\
    TENER & 93.01 & 72.82  & 58.39 & 95.25 \\
    ERNIE & 94.82 & 77.65  & 67.96& 95.08 \\
    \midrule
    Lattice-LSTM & 93.18 & 73.88  & 58.79 & 94.46 \\
    FLAT+BERT & 96.09 & 81.82  & 68.55 & 95.86\\
    LEBERT & 95.7  & 82.08  & 70.75 & 96.08\\
    DyLex+BERT & 96.49 & 81.48  & 71.12 & 95.99 \\
    \midrule
    BERT Baseline & 95.91 & 81.54  & 70.10 & 95.78\\
    TURNER$_{\textrm{MC}}$ & \textbf{96.85} & 83.56  & 70.78 & \textbf{96.36} \\
    TURNER$_{\textrm{Top-}K}$ & 96.54 & \textbf{83.91}  &\textbf{71.22} & \textbf{96.36} \\
    \bottomrule
    \end{tabular}
    }
  
  \caption{The overall performance of TURNER. "BERT Baseline" is the base NER model we utilize in the 1-st stage. "TURNER$_{MC}$" and "TURNER$_{Top-K}$" denote TURNER sampling uncertainty with MC Dropout or Top-K label sequences, respectively.
  }
  \label{tab:main}
\end{table}

%% file: content/ablation.tex
\begin{table}[htbp]
\centering
\scalebox{0.88}{
    \begin{tabular}{lcccc}
    \toprule
  & \textbf{MSRA}  & \textbf{Ontonotes} & \textbf{Weibo} & \textbf{Resume} \\
  \midrule
    % Base NER Model & 95.91 & 81.54  & 70.1 & 95.78\\
    TURNER$_{\textrm{MC}}$ & \textbf{96.85} & \textbf{83.56}  & \textbf{70.78} & \textbf{96.36} \\
    \quad w/o retrieve & 96.33 & 81.93 & 70.15 & 96.11 \\
    \quad w/o uncertainty & 95.85 & 82.45 & 70.27 & 95.87\\
    \quad w/o LE & 96.70 & 82.72 & 69.8 & 96.0 \\
    \quad w/o WL & \textbf{96.85} & 82.60 & \textbf{70.78} & 96.30 \\
    \bottomrule
    \end{tabular}
    }
  
  \caption{Results of the ablation study.}
  \label{tab:ablation}
\end{table}

%% file: content/offline.tex
\begin{table}[htbp]
\centering
\scalebox{0.85}{
    \begin{tabular}{lcccc}
    \toprule
  & \textbf{MSRA}  & \textbf{Ontonotes} & \textbf{Weibo} & \textbf{Resume} \\
    \midrule
    BERT Baseline & 95.91 & 81.54  & 70.10 & 95.78\\
    TURNER$_{\textrm{MC}}$\\
    \quad w/ search engine & 96.85 & 83.56  & 70.78 & 96.36 \\
    \quad w/ offline KG & 96.59 & 82.73  & 70.37 & 96.17 \\
    \bottomrule
    \end{tabular}
    }
  
  \caption{The performance of TURNER with offline retrieval.}
  \label{tab:offline}
\end{table}

%% file: content/compare.tex
% Table generated by Excel2LaTeX from sheet 'Sheet4'
\begin{table}[htbp]
  \centering
  \scalebox{0.9}{
    \begin{tabular}{lcccc}
    \toprule
          & \multicolumn{2}{c}{\textbf{MSRA}} & \multicolumn{2}{c}{\textbf{Ontonotes}} \\
          & MC    & Top-$K$ & MC    & Top-$K$ \\
    \midrule
    SAR & 0.013 & 0.096 & 0.054 & 0.416 \\
    VSR & 0.003 & 0.009 & 0.018 & 0.071 \\
    Avg Retrieval & 0.089 & 0.162 & 0.345 & 0.764 \\
    Oracle F1 & 98.09 & 98.41 & 88.21 & 92.3 \\
    \bottomrule
    \end{tabular}
    }
    \caption{The statistical results of two uncertainty sampling methods. "Avg Retrieval" denotes the average retrieval times for each sample.}
  \label{tab:compare}
\end{table}

%% file: content/appendix.tex
\appendix

\section{Hyperparameters}
\label{sec:hyper}
\input{content/hyperparameter}

\section{Uncertainty Sampling Results}
\label{sec:uresult}
\input{content/uresult}

\section{Case Study}
\input{content/case_study}
\label{sec:case}
In order to better illustrate the effectiveness of retrieved knowledge, we give a case of retrieved knowledge in Table~\ref{tab:case}.

\clearpage
\section{Analysis on Uncertainty Sampling Methods}

\label{sec:analysis}
We give an investigation about the influencing factors of uncertainty sampling using the testset of MSRA, and the results are shown in Table~\ref{tab:mcanalysis}.
For all influencing factors, the performance shows a trend of first rising and then falling, indicating that there are local optimal choices.
\input{content/MCanalysis}

For MC Dropout, as the dropout rate increases, the model's capacity decreases significantly and more noisy candidates are generated.
Therefore, with enough computational resources, finding the optimal forward times is a more general solution.
The Top-K method generates more valuable candidates as well as more noisy candidates. 

%% file: content/hyperparameter.tex
\renewcommand\arraystretch{1.2}
 \begin{table}[htb]
%  \scriptsize
\scalebox{0.85}{
  \centering
  \tabcolsep0.11 in
    \begin{tabular}{lcc}
    \toprule
    & Baseline Model & Knowledge Fusion Model \\
    \midrule
    Epochs& 20 & 10\\
    Batch Size & 32 & 32 \\
    Weight Decay & 0 & 0\\
    Dropout & 0.1 & 0.1\\
    Learning Rate & 2e-5 & \{2e-5,3e-5\}\\
    Optimizer & AdamW & AdamW\\
    Warm Up Ratio & 0.1 & 0.1\\
    Max Seq\_Len & 128 & 512\\
    $\alpha$ & - & \{0.1,1\} \\
    \bottomrule
    \end{tabular}
}
\caption{The hyperparameters we used for training. Other hyperparameters are the same as BERT. For the baseline model, we use same hyperparameters in all datasets. For the Knowledge Fusion Model, the search range is the same in all datasets.}
\label{tab:topkres}
\end{table}

%% file: content/uresult.tex
% Table generated by Excel2LaTeX from sheet 'Sheet3'
\begin{table}[hbp]
  \centering
  \scalebox{0.85}{
    \begin{tabular}{ccccc}
    \toprule
          & \textbf{Type} & \textbf{Train} & \textbf{Validation} & \textbf{Test} \\
          \midrule
    \multicolumn{5}{c}{MC Dropout} \\
    \midrule
    \multirow{2}[0]{*}{\textbf{MSRA}} & Size$_{U}$  & 10748 & -     & 335 \\
          & Num$_{UC}$   & 13748 & -     & 388 \\
          \midrule
    \multirow{2}[0]{*}{\textbf{Ontonotes}} & Size$_{U}$  & 4471  & 1133   & 1229 \\
          & Num$_{UC}$   & 6403  & 1464  & 1501 \\
          \midrule
    \multirow{2}[0]{*}{\textbf{Weibo}} & Size$_{U}$  & 614   & 86    & 93 \\
          & Num$_{UC}$   & 1027  & 133   & 137 \\
          \midrule
    \multirow{2}[0]{*}{\textbf{Resume}} & Size$_{U}$  & 1309  & 50    & 38 \\
          & Num$_{UC}$  & 1798  & 53    & 53 \\
          \midrule
    \multicolumn{5}{c}{Top-$K$ Label Sequences} \\
    \midrule
    \multirow{2}[0]{*}{\textbf{MSRA}} & Size$_{U}$  & - & -     & 615 \\
          & Num$_{UC}$   & - & -     & 706 \\
          \midrule
    \multirow{2}[0]{*}{\textbf{Ontonotes}} & Size$_{U}$  & -  & 2225  & 2383 \\
          & Num$_{UC}$   & - & 3048  & 3322 \\
          \midrule
    \multirow{2}[0]{*}{\textbf{Weibo}} & Size$_{U}$  & -   & 111   & 125 \\
          & Num$_{UC}$   & -  & 155   & 165 \\
          \midrule
    \multirow{2}[0]{*}{\textbf{Resume}} & Size$_{U}$  & -  & 48    & 44 \\
          & Num$_{UC}$   & -  & 50    & 50 \\
          \bottomrule
    \end{tabular}
    }
    \caption{Statistical results of two uncertainty sampling methods. "Size$_{U}$" denotes the number of samples that contain at least one uncertain component. "Num$_{UC}$" denotes the total number of uncertain components in all samples.}
  \label{tab:uresult}
\end{table}%

%% file: content/case_study.tex
\begin{CJK}{UTF8}{gbsn}
\begin{table}[b]
\centering
\scalebox{1.0}{
\begin{tabular}{p{3in}}
\toprule
\textbf{Original Text} \\
\midrule
实践越来越证明,“\textcolor{red}{亚太经合组织}方式”符合本地区的实际,有利于各成员的不同权益和要求得到较好的平衡,有助于发挥各个成员的能力,促进共同发展。\\
\midrule
\textbf{Uncertain Component} \\
\midrule
亚太经合组织 \\
\midrule
\textbf{Knowledge w/ Search Engine} \\
\midrule
亚太经合组织-百度百科:简介：亚太经合组织一般指亚太经济合作组织。亚太经济合作组织（英文：asia-pacificeco\textcolor{red}{\textbf{|}}亚太经合组织-asia-pacificeconomiccooperat...:apecministers'meeting08-09november2021vir\textcolor{red}{\textbf{|}}亚太经合组织超额实现悉尼林业目标\_政务\_澎湃新闻-thepaper:11月9日,亚太经合组织部长级会议联合发布部长声明,对亚太区域森林面积增长2790万公顷,如期实现悉\textcolor{red}{\textbf{|}}习近平在亚太经合组织工商领导人峰会上的主旨演讲(全文):坚持可持续发展共建亚太命运共同体在亚太经合组织工商领导人峰会上的主旨演讲（2021年11月1\textcolor{red}{\textbf{|}}亚太经合组织概况中华人民共和国外交部:【成立】亚太经合组织(asia-pacificeconomiccooperation,ape\textcolor{red}{\textbf{|}}习近平在亚太经合组织第二十八次领导人非正式会议上的讲话...:很高兴同大家见面。首先,我感谢阿德恩总理和新西兰政府为本次会议作出的努力。本次会议以推动疫后经济复\textcolor{red}{\textbf{|}}增加2650万公顷!中国为亚太地区森林增长贡献巨大\textcolor{red}{\textbf{|}}亚太经合组织超额实现悉尼林业目标\textcolor{red}{\textbf{|}}澳媒社论:亚太经济合作攸关全球进步\textcolor{red}{\textbf{|}}悉尼林业目标超额实现,中国森林面积13年增2650...\textcolor{red}{\textbf{|}}携手开创亚太经济合作新篇章\\
\midrule
\textbf{Knowledge w/ Knowledge Graph}\\
\midrule
亚太经合组织第二十二次领导人非正式会议宣言。北京纲领北京纲领：构建融合、创新、互联的亚太亚太经合组织第二十二次领导人非正式会议宣言我们，亚太经合组织各成员领导人聚首北京雁栖湖畔，举行亚太经合组织第二十二次领导人非正式会议。。中文名:亚太经合组织第二十二次首届亚太经合组织林业部长级会议。\textcolor{red}{\textbf{|}}首届亚太经合组织林业部长级会议，在北京人民大会堂开幕。。中文名:首届亚太经合组织林业部长级会议。类别:会议。地点:亚太经合组织。类型:林业部长。标签:社会事件。歧义关系:首届亚太经合组织林业部长级会议。歧义权重:1028。\textcolor{red}{\textbf{|}}亚太经合组织领导人非正式会议。亚太经合组织领导人非正式会议，是亚太经合组织高级别的会议。。中文名:亚太经合组织领导人非正式会议。外文名:theannualapececonomicleaders;meetings。成立时间:1992年4月。首届\\
\bottomrule
\end{tabular}
}
\caption{An example of retrieved knowledge.}
\label{tab:case}
\end{table}
\end{CJK}

%% file: content/MCanalysis.tex
\begin{table}[htbp]
  \centering
    \begin{tabular}{lcccc}
    \toprule
    \multicolumn{5}{c}{MC Dropout, Candidate Num = 8} \\
    \midrule
    Drop Rate & 0.05  & 0.1   & 0.2   & 0.4 \\
    SAR & 0.008 & 0.013 & 0.031 & 0.221 \\
    VSR & 0.002 & 0.003 & 0.005 & 0.002 \\
    F1 Score & 96.74 & 96.85 & 96.66 & 96.29 \\
    \midrule
    \multicolumn{5}{c}{MC Dropout, Drop Rate = 0.1} \\
    \midrule
    Candidate Num & 2     & 4     & 8     & 16 \\
    SAR & 0.023 & 0.018 & 0.013 & 0.009 \\
    VSR & 0.006 & 0.005  & 0.003 & 0.002 \\
    F1 Score & 96.56 & 96.79 & 96.85 & 96.75 \\
    \midrule
    \multicolumn{5}{c}{Top-$K$ Method} \\
    \midrule
    Candidate Num & 1     & 2     & 4     & 8 \\
    SAR & 0.118 & 0.104 & 0.096 & 0.088 \\
    VSR & 0.019 & 0.014 & 0.009 & 0.006 \\
    F1 Score & 96.42 & 96.50 & 96.54 & 96.46 \\
    \bottomrule
    \end{tabular}%
  \caption{The results on MSRA datasets of our investigation.}
    \label{tab:mcanalysis}
\end{table}%

%% file: ijcai22.bbl
\begin{thebibliography}{}

\bibitem[\protect\citeauthoryear{Brown and Golod}{2010}]{brown2010decoding}
Daniel~G Brown and Daniil Golod.
\newblock Decoding hmms using the k best paths: algorithms and applications.
\newblock {\em BMC bioinformatics}, 11(1):1--7, 2010.

\bibitem[\protect\citeauthoryear{Gal and
  Ghahramani}{2016}]{YarinGal2016DropoutAA}
Yarin Gal and Zoubin Ghahramani.
\newblock Dropout as a bayesian approximation: representing model uncertainty
  in deep learning.
\newblock In {\em International Conference on Machine Learning}, 2016.

\bibitem[\protect\citeauthoryear{Gu \bgroup \em et al.\egroup
  }{2018}]{gu2018search}
Jiatao Gu, Yong Wang, Kyunghyun Cho, and Victor~OK Li.
\newblock Search engine guided neural machine translation.
\newblock In {\em Proceedings of the AAAI Conference on Artificial
  Intelligence}, volume~32, 2018.

\bibitem[\protect\citeauthoryear{Gui \bgroup \em et al.\egroup
  }{2020}]{TaoGui2020UncertaintyAwareLR}
Tao Gui, Jiacheng Ye, Qi~Zhang, Zhengyan Li, Zichu Fei, Yeyun Gong, and
  Xuanjing Huang.
\newblock Uncertainty-aware label refinement for sequence labeling.
\newblock In {\em Empirical Methods in Natural Language Processing}, 2020.

\bibitem[\protect\citeauthoryear{Hashimoto \bgroup \em et al.\egroup
  }{2018}]{TatsunoriBHashimoto2018ARF}
Tatsunori~B. Hashimoto, Kelvin Guu, Yonatan Oren, and Percy Liang.
\newblock A retrieve-and-edit framework for predicting structured outputs.
\newblock In {\em Neural Information Processing Systems}, 2018.

\bibitem[\protect\citeauthoryear{He \bgroup \em et al.\egroup
  }{2020}]{he2020knowledge}
Qizhen He, Liang Wu, Yida Yin, and Heming Cai.
\newblock Knowledge-graph augmented word representations for named entity
  recognition.
\newblock In {\em Proceedings of the AAAI Conference on Artificial
  Intelligence}, volume~34, pages 7919--7926, 2020.

\bibitem[\protect\citeauthoryear{Houlsby \bgroup \em et al.\egroup
  }{2019}]{NeilHoulsby2019ParameterEfficientTL}
Neil Houlsby, Andrei Giurgiu, Stanisław Jastrzębski, Bruna~Halila Morrone,
  Quentin de~Laroussilhe, Andrea Gesmundo, Mona Attariyan, and Sylvain Gelly.
\newblock Parameter-efficient transfer learning for nlp.
\newblock In {\em International Conference on Machine Learning}, 2019.

\bibitem[\protect\citeauthoryear{Huang \bgroup \em et al.\egroup
  }{2015}]{huang2015bidirectional}
Zhiheng Huang, Wei Xu, and Kai Yu.
\newblock Bidirectional lstm-crf models for sequence tagging.
\newblock {\em arXiv preprint arXiv:1508.01991}, 2015.

\bibitem[\protect\citeauthoryear{Levow}{2006}]{levow2006third}
Gina-Anne Levow.
\newblock The third international chinese language processing bakeoff: Word
  segmentation and named entity recognition.
\newblock In {\em Proceedings of the Fifth SIGHAN Workshop on Chinese Language
  Processing}, pages 108--117, 2006.

\bibitem[\protect\citeauthoryear{Li \bgroup \em et al.\egroup
  }{2020}]{li2020flat}
Xiaonan Li, Hang Yan, Xipeng Qiu, and Xuan-Jing Huang.
\newblock Flat: Chinese ner using flat-lattice transformer.
\newblock In {\em Proceedings of the 58th Annual Meeting of the Association for
  Computational Linguistics}, pages 6836--6842, 2020.

\bibitem[\protect\citeauthoryear{Liu \bgroup \em et al.\egroup
  }{2021}]{liu-etal-2021-lexicon}
Wei Liu, Xiyan Fu, Yue Zhang, and Wenming Xiao.
\newblock Lexicon enhanced {C}hinese sequence labeling using {BERT} adapter.
\newblock In {\em Proceedings of the 59th Annual Meeting of the Association for
  Computational Linguistics and the 11th International Joint Conference on
  Natural Language Processing (Volume 1: Long Papers)}, pages 5847--5858,
  Online, August 2021. Association for Computational Linguistics.

\bibitem[\protect\citeauthoryear{Ma and Hovy}{2016}]{XuezheMa2016EndtoendSL}
Xuezhe Ma and Eduard Hovy.
\newblock End-to-end sequence labeling via bi-directional lstm-cnns-crf.
\newblock In {\em Meeting of the Association for Computational Linguistics},
  2016.

\bibitem[\protect\citeauthoryear{Peng and Dredze}{2015}]{peng2015named}
Nanyun Peng and Mark Dredze.
\newblock Named entity recognition for chinese social media with jointly
  trained embeddings.
\newblock In {\em Proceedings of the 2015 Conference on Empirical Methods in
  Natural Language Processing}, pages 548--554, 2015.

\bibitem[\protect\citeauthoryear{Qiu \bgroup \em et al.\egroup
  }{2014}]{XipengQiu2014AutomaticCE}
Xipeng Qiu, Chaochao Huang, and Xuanjing Huang.
\newblock Automatic corpus expansion for chinese word segmentation by
  exploiting the redundancy of web information.
\newblock In {\em International Conference on Computational Linguistics}, 2014.

\bibitem[\protect\citeauthoryear{Sun \bgroup \em et al.\egroup
  }{2019}]{sun2019ernie}
Yu~Sun, Shuohuan Wang, Yukun Li, Shikun Feng, Xuyi Chen, Han Zhang, Xin Tian,
  Danxiang Zhu, Hao Tian, and Hua Wu.
\newblock Ernie: Enhanced representation through knowledge integration.
\newblock {\em arXiv preprint arXiv:1904.09223}, 2019.

\bibitem[\protect\citeauthoryear{Torisawa and
  others}{2007}]{torisawa2007exploiting}
Kentaro Torisawa et~al.
\newblock Exploiting wikipedia as external knowledge for named entity
  recognition.
\newblock In {\em Proceedings of the 2007 joint conference on empirical methods
  in natural language processing and computational natural language learning
  (EMNLP-CoNLL)}, pages 698--707, 2007.

\bibitem[\protect\citeauthoryear{Wang \bgroup \em et al.\egroup
  }{2021a}]{wang2021dylex}
Baojun Wang, Zhao Zhang, Kun Xu, Guang-Yuan Hao, Yuyang Zhang, Lifeng Shang,
  Linlin Li, Xiao Chen, Xin Jiang, and Qun Liu.
\newblock Dylex: Incorporating dynamic lexicons into bert for sequence
  labeling.
\newblock In {\em Proceedings of the 2021 Conference on Empirical Methods in
  Natural Language Processing}, pages 2679--2693, 2021.

\bibitem[\protect\citeauthoryear{Wang \bgroup \em et al.\egroup
  }{2021b}]{wang-etal-2021-improving}
Xinyu Wang, Yong Jiang, Nguyen Bach, Tao Wang, Zhongqiang Huang, Fei Huang, and
  Kewei Tu.
\newblock Improving named entity recognition by external context retrieving and
  cooperative learning.
\newblock In {\em Proceedings of the 59th Annual Meeting of the Association for
  Computational Linguistics and the 11th International Joint Conference on
  Natural Language Processing (Volume 1: Long Papers)}, pages 1800--1812,
  Online, August 2021. Association for Computational Linguistics.

\bibitem[\protect\citeauthoryear{Weischedel \bgroup \em et al.\egroup
  }{2011}]{weischedel2011ontonotes}
Ralph Weischedel, Sameer Pradhan, Lance Ramshaw, Martha Palmer, Nianwen Xue,
  Mitchell Marcus, Ann Taylor, Craig Greenberg, Eduard Hovy, Robert Belvin,
  et~al.
\newblock Ontonotes release 4.0.
\newblock {\em LDC2011T03, Philadelphia, Penn.: Linguistic Data Consortium},
  2011.

\bibitem[\protect\citeauthoryear{Yan \bgroup \em et al.\egroup
  }{2019}]{yan2019tener}
Hang Yan, Bocao Deng, Xiaonan Li, and Xipeng Qiu.
\newblock Tener: adapting transformer encoder for named entity recognition.
\newblock {\em arXiv preprint arXiv:1911.04474}, 2019.

\bibitem[\protect\citeauthoryear{Yang \bgroup \em et al.\egroup
  }{2017}]{JieYang2017NeuralRF}
Jie Yang, Yue Zhang, and Fei Dong.
\newblock Neural reranking for named entity recognition.
\newblock In {\em Recent Advances in Natural Language Processing}, 2017.

\bibitem[\protect\citeauthoryear{Zhang and Yang}{2018}]{YueZhang2018ChineseNU}
Yue Zhang and Jie Yang.
\newblock Chinese ner using lattice lstm.
\newblock In {\em Meeting of the Association for Computational Linguistics},
  2018.

\end{thebibliography}
